\documentclass[a4paper]{svproc}
\usepackage{url}

\usepackage[backend=biber,
            url=false,
            isbn=false,
            doi=false,
            backref=false,
            natbib=true,
            mincitenames=1,
            maxcitenames=1,
            citestyle=numeric-comp,
            sorting=none,
            block=none]{biblatex}

\AtBeginBibliography{\setcounter{maxnames}{500}}
            
\usepackage{multicol}
\usepackage[bookmarks=true]{hyperref}
\usepackage{adjustbox}
\usepackage{amsmath,amssymb,amsfonts}

\newcommand{\R}{\mathbb{R}}

\usepackage{multirow}
\usepackage{graphicx}
\usepackage{color}
\usepackage{caption}
\usepackage{booktabs}
\usepackage{bm}
\usepackage{verbatim}
\usepackage{gensymb}
\usepackage{threeparttable}
\usepackage{siunitx}
\graphicspath{{figures/}}

\renewcommand{\remark}[3]{} 

\definecolor{amber}{rgb}{1.0, 0.49, 0.0}

\definecolor{britishracinggreen}{rgb}{0.23, 0.53, 0.19}

\definecolor{blueviolet}{rgb}{0.54, 0.17, 0.89}

\definecolor{navy}{rgb}{0,0,0.5}

\begin{document}
\sloppy

\mainmatter              
\title{Efficiently Learning Single-Arm Fling Motions to Smooth Garments}
\titlerunning{Efficiently Learning Single-Arm Fling Motions to Smooth Garments}  
%
\author{Lawrence Yunliang Chen*\inst{1} \and Huang Huang*\inst{1} \and Ellen Novoseller\inst{1} \and  Daniel Seita\inst{2} \and \\ Jeffrey Ichnowski\inst{1} \and  Michael Laskey\inst{3} \and Richard Cheng\inst{4} \and Thomas Kollar\inst{4}  \and  Ken Goldberg \inst{1} {\scriptsize * equal contribution}}
\authorrunning{Chen*, Huang*, et al.} 
%
\tocauthor{Lawrence Yunliang Chen, Huang Huang, Ellen Novoseller, Daniel Seita, Jeffrey Ichnowski,  Michael Laskey, Richard Cheng, Thomas Kollar,  Ken Goldberg}
\institute{University of California, Berkeley, Berkeley, CA 94720, USA$^1$\thanks{$^{1}$The AUTOLab at UC Berkeley (automation.berkeley.edu).},\\
\and
Carnegie Mellon University, Pittsburgh, PA 15213, USA\\
\and
Electric Sheep Robotics, San Francisco, CA 94131, USA\\
\and
Toyota Research Institute, Los Altos, CA 94022, USA\\
\email{ \{yunliang.chen, huangr\} @berkeley.edu}
}

\maketitle
\begin{abstract} 
Recent work has shown that 2-arm ``fling'' motions can be effective for garment smoothing. We consider single-arm fling motions. Unlike 2-arm fling motions, which require little robot trajectory parameter tuning, single-arm fling motions are very sensitive to trajectory parameters. We consider a single 6-DOF robot arm that learns fling trajectories to achieve high garment coverage. Given a garment grasp point, the robot explores different parameterized fling trajectories in physical experiments. To improve learning efficiency, we propose a coarse-to-fine learning method that first uses a multi-armed bandit (MAB) framework to efficiently find a candidate fling action, which it then refines via a continuous optimization method. Further, we propose novel training and execution-time stopping criteria based on fling outcome uncertainty; the training-time stopping criterion increases data efficiency while the execution-time stopping criteria leverage repeated fling actions to increase performance. Compared to baselines, the proposed method significantly accelerates learning. Moreover, with prior experience on similar garments collected through self-supervision, the MAB learning time for a new garment is reduced by up to 87\%. We evaluate on 36 real garments: towels, T-shirts, long-sleeve shirts, dresses, sweat pants, and jeans. Results suggest that using prior experience, a robot requires under 30 minutes to learn a fling action for a novel garment that achieves 60--94\% coverage. Supplementary material can be found at \url{https://sites.google.com/view/single-arm-fling}.
\end{abstract}

\section{Introduction}
Garment smoothing~\cite{cloth_icra_2015} is useful in daily life, industry, and clothing stores as a first step toward garment folding~\cite{maitin2010cloth,laundry2012,balaguer2011combining}, and hanging, as well as assistive dressing~\cite{assistive_gym_2020,deep_dressing_2018}.
However, almost all prior research in robot garment manipulation has focused on quasistatic pick-and-place motions~\cite{VCD_cloth,fabricflownet,lerrel_2020,yan_fabrics_latent_2020,latent_space_roadmap_2020,seita_fabrics_2020,seita_bags_2021}, which are relatively slow and limited by the robot workspace. 
Yet, robots can perform dynamic actions such as casting, throwing, vaulting~\cite{harry_rope_2021, PRC_2022, ha2021flingbot}, and air blowing~\cite{xu2022dextairity} to efficiently place objects such as ropes, garments, and bags in desired configurations, even when much of the object lies outside the robot workspace. 
Dynamic manipulation of deformable objects is challenging~\cite{manip_deformable_survey_2018,lynch1999dynamic,zeng_tossing_2019} as such objects are partially observable, have infinite-dimensional configuration spaces, and have uncertain and 
complex dynamics due to deformability, elasticity, and friction. 
To address these modeling challenges, this work explores a learning-based approach for garment smoothing via high-speed dynamic actions that use inertia to move garments beyond the robot workspace. 

\begin{figure}[t]
  \centering
    \includegraphics[height = 5cm]{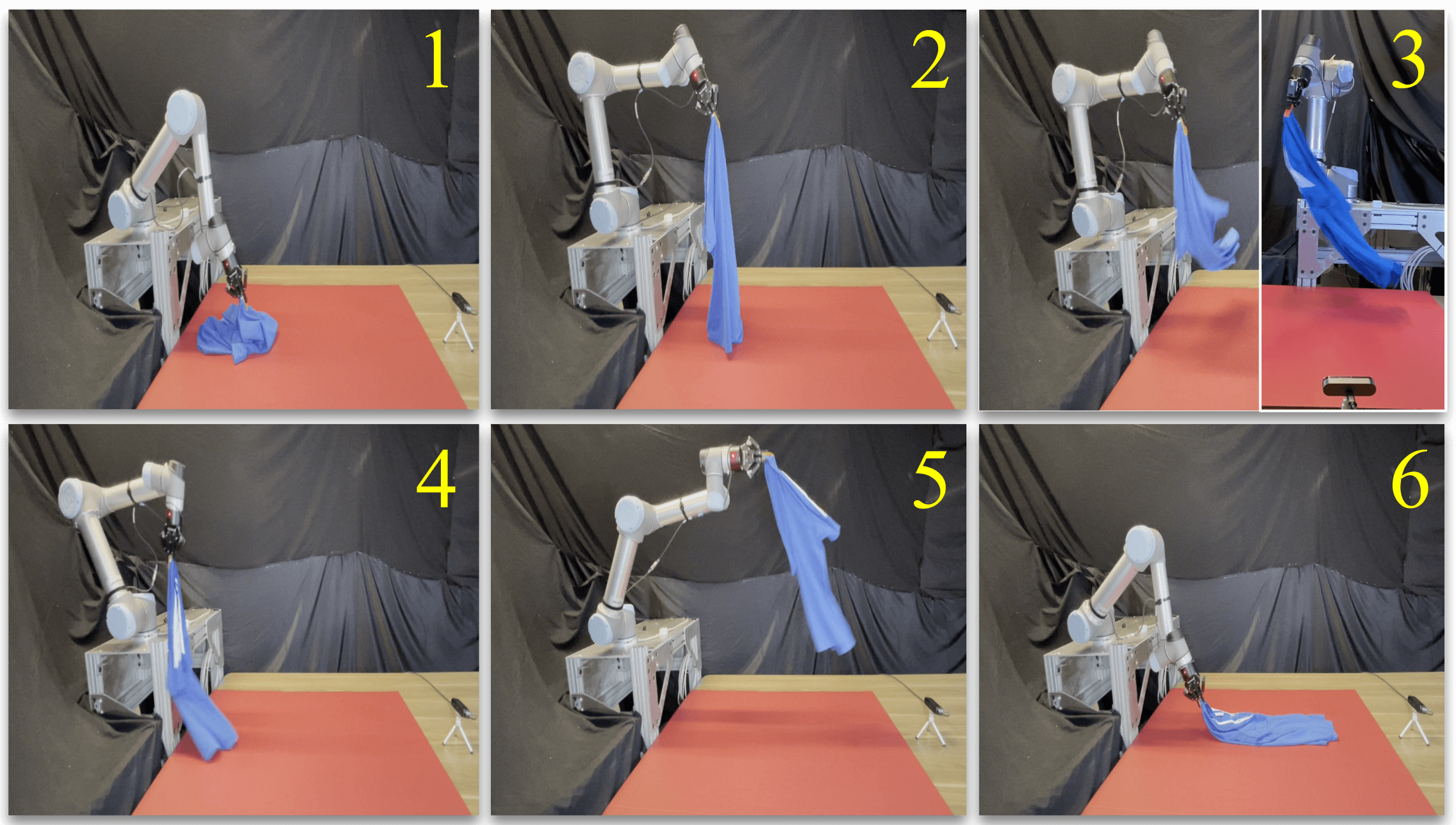}
    \caption{\textbf{Single-arm dynamic garment flinging.}  A UR5 robot flings a blue T-shirt. \textbf{Top row}: the robot holding the garment (1), lifts (2) and shakes it vertically (3 left) and horizontally (3 right) to loosen the garment using gravity. \textbf{Bottom row}: The robot flings the garment using the learned parameterized trajectory (Section~\ref{sec:fling}) to smooth it on the surface. This cycle repeats until a stopping criterion is satisfied.}
    \label{fig:teaser}
  \vspace{-20pt}
\end{figure}


In a pioneering paper, Ha and Song~\cite{ha2021flingbot} proposed FlingBot, which used dynamic 2-arm ``fling'' motions to smooth garments. Using scripted motions at constant speed and air resistance, they achieved over 90\% coverage for T-shirts. However, dual-arm manipulators may be too costly and bulky for home use. Many mobile manipulators, such as the HSR~\cite{hsr2013} and the Fetch~\cite{fetch}, only have 1 arm. In this work, we show that a robot can use 1-arm fling motions to achieve a coverage of 60--94\,\% for 6 garment types. 

Single-arm flings are more difficult than 2-arm flings, since they cannot straighten one garment edge prior to flinging. 
We propose a single-arm dynamic smoothing pipeline with novel parameterized primitives including shaking and flinging, and we propose an efficient method to learn these parameters. We begin with the robot gripping a fixed location on a crumpled garment. The robot lifts the garment and shakes it horizontally and vertically to loosen and reduce wrinkles. The robot then executes a learned fling motion. It retries the fling 
until a stopping rule determines that the coverage is sufficient. We propose a self-supervised learning method that uses the multi-armed bandit (MAB) framework with Thompson Sampling (TS) to efficiently find a good candidate fling motion and then refines the fling parameters using the cross-entropy method (CEM). 

The robot learns through real-world experiments.
While prior work~\cite{PRC_2022,ha2021flingbot} shows that simulation can facilitate learning for robot manipulation,
state-of-the-art deformable object simulators have a large reality gap~\cite{reality_gap_1995}, and dynamic garment flings may require modeling additional effects such as material properties and air resistance.
Consequently, we directly optimize flinging through real-world online learning and improve data efficiency by introducing a stopping rule for online training to threshold the expected performance improvement. 

Unlike quasistatic actions, each fling results in significant variance in coverage due to the inherent \emph{aleatoric} uncertainty, i.e., non-repeatable variation in outcomes. 
Thus, we propose a \emph{second} stopping rule for execution time that stops the robot flinging when the coverage exceeds a learned threshold or 
maximum number of attempts. This stopping rule differs from the training-time stopping rule, which increases data efficiency.

This paper makes 4 contributions:
\begin{enumerate}
    \item A formulation of single-arm garment flinging and a robot system that uses self-supervised learning in real to efficiently smooth garments.
    \item An online learning framework that operates under large aleatoric uncertainty and enables the robot to quickly adapt to novel garments by leveraging a learned garment category prior and by optimizing fling parameters in a coarse-to-fine manner.
    \item A training-time stopping rule based on epistemic uncertainty and execution-time stopping rules that exploit the aleatoric uncertainty of dynamic actions to increase expected execution-time fling coverage.
    \item Experiments with 36 real garments (30 for training, 6 for testing) 
    suggesting that the robot can learn effective single-arm fling actions achieving average coverage of 60--94\,\% on novel garments in 30 minutes.
\end{enumerate}

\section{Related Work}
Deformable object manipulation has a long history in robotics research~\cite{grasp_centered_survey_2019,manip_deformable_survey_2018,2021_survey_defs}. This paper focuses specifically on manipulation of garments.
Much prior research on cloth manipulation uses geometric methods for perception and planning~\cite{manipulation_cloth_liu_2016,regrasp_unfold_2015,unfolding_rf_2014,kita_2009_icra,kita_2009_iros}. 
While these works show promising results, they make strong assumptions on garment configuration. To simplify planning, many of these works ``funnel'' garments into known states, for instance iteratively re-grasping the lowest hanging corner or using gravity for vertical smoothing~\cite{maitin2010cloth,cusumano2011bringing}. 
\vspace{-10pt}
\subsubsection{Learning Garment Manipulation}
The complexities of crumpled garments make representing their states difficult~\cite{chi2021garmentnets}.
Researchers have thus bypassed precise state estimation by using data-driven methods such as imitation learning~\cite{argall2009survey} and reinforcement learning. 
Prior learning-based approaches for garments mainly use quasistatic actions (e.g., pick-and-place) for smoothing~\cite{seita-bedmaking,seita_fabrics_2020,seita_bags_2021,lerrel_2020,VCD_cloth,latent_space_roadmap_2020,yan_fabrics_latent_2020} and folding~\cite{fabric_vsf_2020,folding_fabric_fcn_2020,fabricflownet}. While attaining promising results, they may require many short actions and are limited by the robot's reach. 
In contrast, 
dynamic actions can manipulate garments with fewer actions, including out-of-reach portions.
\vspace{-20pt}
\subsubsection{Dynamic Manipulation} \label{RW:dynamic_manipulation}
In dynamic manipulation, robots leverage high-speed actions~\cite{lynch1999dynamic} such as tossing~\cite{zeng_tossing_2019}, swinging~\cite{swingbot_2020}, and blowing~\cite{xu2022dextairity}. There has been considerable progress on deformable dynamic manipulation, for instance high-speed cable knotting~\cite{dynamic_knotting_2010,high_speed_knotting_2013} and learning-based approaches for high-speed manipulations of cables with a fixed endpoint~\cite{harry_rope_2021} and a free endpoint~\cite{PRC_2022,zimmermanndynamic,chi2022irp}. Work on dynamic \emph{garment} manipulation includes~\citet{rishabh_2020} and \citet{Closing_Sim2Real_Dynamic_Fabric_2021}, which both leverage simulators, with the latter transferring to real; in contrast, we do not use simulation due to the reality gap and challenges in Sim2Real transfer~\cite{reality_gap_1995}.
While there have been great strides in deformable simulation with quasistatic actions, simulating the dynamic movement of deformable objects including the effect of aerodynamics remains challenging~\cite{rss_workshop}. 

Recently, Ha and Song presented FlingBot, which learns to perform a fixed-parameter dynamic fling motion to smooth garments using two UR5 arms. FlingBot is initially trained in SoftGym simulation~\cite{corl2020softgym} and fine-tuned in real, and learns grasp points for both arms. 
The results suggest significant efficiency benefits in attaining high coverage compared to quasistatic pick-and-place actions. 
In contrast, we consider using a \emph{single-arm} manipulator and directly learn actions in real. While it is much harder to smooth a garment with only one arm---since two can effectively stretch one side of the garment so that the fling must only straighten the other side---single-arm fling motions can reduce the hardware cost or increase efficiency via multiple arms running in parallel. 
\vspace{-10pt}
\section{Problem Statement}



\subsubsection{Workspace Assumptions}
\label{subsec:workspace_definition}
We define a Cartesian coordinate frame containing a single-arm UR5 robot and a manipulation surface parallel to the $xy$-plane (Fig.~\ref{fig:traj}), 
faced by an overhead RGB camera. We estimate garment coverage via the overhead RGB images with a manually-measured HSV threshold for each garment, and using the OpenCV package. As in some prior work on dynamic robot manipulation~\cite{harry_rope_2021,PRC_2022,Closing_Sim2Real_Dynamic_Fabric_2021}, we assume that the robot gripper holds the garment at a fixed point. 
During experiments, the initial grasp is set by a human.

We select the fixed grasping point by studying the performance of various grasp points with a T-shirt (see Section~\ref{ssec:different_grasp_points}). The \emph{back collar} (Fig.~\ref{fig:result-grasp_point}) yields the most promising results. Thus, in all subsequent experiments, the robot grasps the \emph{back collar} of shirts and dresses. For towels, the robot grasps the center top point, while for jeans and pants, the robot grasps the center of the waistband's backside, as these locations are most analogous to the back collar of a T-shirt.

\vspace{-10pt}
\subsubsection{Task Objective}
The objective is to efficiently learn single-arm robot fling motions that maximally increase the coverage of an arbitrarily-crumpled garment. We assume access to training and testing sets of garments, denoted $G_{\rm train}$ and $G_{\rm test}$. The fling motions are learned on the training set. Once training concludes, the robot executes the learned fling motions at execution time. During execution, repeated flinging is allowed up to a repetition limit. 


We evaluate performance via the average garment coverage achieved by the learned fling action, similar to prior work on dynamic garment manipulation~\cite{ha2021flingbot} and garment manipulation more generally~\cite{seita_fabrics_2020,seita_bags_2021}. 
We also aim to reduce the number of flinging trials  
needed during training, where each trial corresponds to the execution of a fling cycle as described in Section~\ref{sec:fling}. 
For each test garment, we record the number of flings the algorithm takes to learn the best fling action. 

\section{Motion Primitives and Pipeline}\label{sec:fling}
For each selected fling action, we execute one or more fling cycles to smooth the garment. 
A fling cycle consists of a reset motion, two shaking motions, and a dynamic arm motion. The gripper is closed during this process.
\vspace{-10pt}
\subsubsection{Reset Procedure}\label{ssec:reset}
To learn effective fling motions for different initial garment states, the self-supervised learning procedure performs a \emph{reset} motion at the start of each fling cycle to randomize the garment's initial state. 
The reset has two parts: 1) the robot crumples the garment by lifting to a fixed joint configuration $\mathbf{q}_{\rm up}$ and then placing it down at $\mathbf{q}_{\rm down}$; and 2) the robot returns to $\mathbf{q}_{\rm up}$ to help smooth the garment using gravity~\cite{maitin2010cloth}. 
While the garment is not in an identical state after each reset (see website), and the folds can assume different shapes and directions, the reset places the garment in a stable position with its center of mass below the gripper, increasing repeatability of subsequent dynamic actions.
\vspace{-15pt}
\subsubsection{Shaking Motions}
Different garment layers can often stick together due to friction, making it more difficult for a fling action to expand the garment. 
In addition, garments often exhibit twists while hanging, resulting in correlated initial states between trials. 
Thus, immediately following the reset, we use two shaking motions---horizontal and vertical, shown in Fig.~\ref{fig:teaser}---to loosen parts of garments that may be stuck together and to reduce inter-trial correlation.
Both shaking motions have a predefined amplitude, and each is repeated 3 times.  The period $T$ is chosen to be as brief as possible subject to robot motion control constraints.
Vertical shaking plays the more important role of loosening the garments, while horizontal shaking mainly helps to reduce any twists. 



\vspace{-10pt}
\subsubsection{Fling Parameterization}
\label{ssec:fling-parameterization}


Inspired by observing humans performing single-arm flings, we parameterize the fling motion using a trajectory as shown in Fig.~\ref{fig:traj}. We control the robot's movement through both the spatial position and angle of joint 3. Joint 3 starts at point 1 (P1), pulls back to point 2 (P2) to create more space for the actual fling motion, accelerates from P2 to point 3 (P3) to perform the fling, and finally lays down the garment by moving to point 4 (P4). All four points lie on the $yz$-plane (i.e., $x=0$). We accelerate learning by reducing the parameter space dimension. During preliminary experiments, varying P2 and P4 did not significantly affect results, and hence we fix those points. 

The fling motion is therefore parameterized by 7 learnable \emph{fling parameters},
$
(v^{23}_{max}, v^{34}_{max},P_{3, y}, P_{3, z}, \theta, v_{\theta}, a_{\theta}) \in \R^7,
$
where $v^{23}_{max}$ and $v^{34}_{max}$ are maximum joint velocities for all joints (Fig.~\ref{fig:traj}) while traveling from P2 to P3 and from P3 to P4, respectively; $\theta$ and $(v_{\theta}$, $a_{\theta})$ are the angle, velocity and acceleration of joint 3 at P3; and $(0, P_{3, y}, P_{3, z})$ gives the $(x,y,z)$-coordinates of P3.
In FlingBot~\cite{ha2021flingbot}, contrastingly, the authors use a universal scripted fling motion for all garments and instead focus on learning the grasp points for their two arms.
We use Ruckig~\cite{berscheid2021jerklimited}, a time-optimal trajectory generation algorithm, to compute a trajectory passing through all waypoints 
with the desired velocities and accelerations at Points 1--4, while limiting velocity, acceleration, and jerk.

\section{Self-Supervised Learning Pipeline for Flinging} \label{sec:learning_procedure}
To accomplish the task objective, we propose a self-supervised learning procedure to learn the fling parameters for a given garment, optionally using prior experience with similar garments to accelerate learning. 
\subsection{Learning Procedure of the Fling Action}
Due to the high aleatoric uncertainty and continuous parameter space, we propose to explore the fling parameter space in a coarse-to-fine manner. 
First, we use a multi-armed bandit (MAB)~\cite{auer2002finite} algorithm on discretized bins to quickly identify the best region of the parameter space.  Then, we fine-tune the parameters with the cross-entropy method (CEM)~\cite{cem_2004}. 
The MAB framework is well-suited for efficiently identifying well-performing actions under the high aleatoric uncertainty in fling outcomes, while CEM does not require assuming a specific form for the underlying distribution.

For MAB, we form a set of actions $\mathcal{A} = \{a_1, \ldots, a_K\}$ by dividing the flinging parameter space into a uniform grid consisting of $K$ grid cells, where each action $a_k \in \mathbb{R}^7$ is a grid cell center 
and parameterizes a flinging motion. Each parameter’s range is defined based on the robot's physical limits and human observation (see Section~\ref{ssec:fling-parameterization}). 
During each learning iteration $t$, the MAB algorithm samples an action $a^{(t)} \in \mathcal{A}$ to execute in the next fling cycle and observes a stochastic reward $r_t \in [0,1]$ measuring post-fling garment coverage, computed using the overhead camera and expressed as a percentage of the garment's full surface area in its fully-smoothed state. 
After $T$ steps, the observed actions and rewards form a dataset, $\mathcal{D}_T := \{(a^{(1)}, r^{(1)}), \ldots, (a^{(T)}, r^{(T)})\}$. Learning continues for a predefined iteration limit or until satisfying a \emph{stopping rule} based on the estimated expected improvement. After the MAB learning procedure, we refine the MAB-chosen action using CEM, a continuous optimization algorithm, where we constrain CEM within the fling parameter grid cell corresponding to the MAB-chosen action. At each iteration, CEM samples a batch of actions from a normal distribution, which are then executed by the robot. The top few actions with the highest coverage (termed the ``elites'') 
are used to update the mean and variance of the CEM sampling distribution for the next CEM iteration. 
\vspace{-10pt}
\subsubsection{Multi-Armed Bandit Algorithm}
We use Thompson Sampling (TS)~\cite{thompson1933likelihood} as the MAB algorithm. TS maintains a Bayesian posterior over the reward of each action $a \in \mathcal{A}$: $p(r \mid a, \mathcal{D}_t)$. We use a Gaussian distribution to model both the prior and posterior reward of each action, and initialize the prior to $\mathcal{N}(0.5, 1)$ for all actions. On each iteration $t$, TS draws a reward sample $\tilde{r}_i \sim p(r \mid a_i, \mathcal{D}_t)$ for each action $a_i \in \mathcal{A}$, $i \in \{1, \ldots, K\}$. Then, TS selects the action corresponding to the highest sampled reward: $a^{(t)} = a_j, j = \text{argmax}_i\{\tilde{r}_i \mid i = 1, \ldots, K\}$. 

\vspace{-10pt}
\subsubsection{Stopping Rule during Learning}\label{sssec:stopping}
Physical robot learning can be expensive due to the high execution time and potential risk of collision, making data efficiency important. During MAB learning, we balance efficiency with performance by stopping online learning when the estimated expected improvement over the highest posterior mean reward of any action falls below a given threshold~\cite{snoek2012practical}. Otherwise, the algorithm continues learning until an iteration limit. We calculate the expected improvement analytically~\cite{snoek2012practical} as follows, where $\phi(\cdot)$ and $\Phi(\cdot)$ are the PDF and CDF of the standard normal distribution, respectively, and $\mu(a_i)$ and $\sigma(a_i)$ are the posterior mean and standard deviation of action $a_i$:
\begin{align*}
    \text{EI}(a_i) = \left(\mu(a_i) - \max_i\{\mu(a_i)\}\right) \Phi(Z) + \sigma(a_i) \phi(Z), \,\,\,
    Z = \frac{\mu(a_i) - \max_i\{\mu(a_i)\}}{\sigma(a_i)}.
\end{align*}

\vspace{-20pt}
\subsection{Adapting to New Garments}\label{ssec:learning-adapt}
To efficiently adapt to novel garments, we propose a learning algorithm that leverages prior experience from previously-seen garments. The algorithm consists of a training stage followed by an online learning stage. During training, we collect data from a training dataset $G_{\rm train}$ of $m$ garments from all categories by executing the MAB algorithm and recording the final converged distributions of each action's coverage after MAB learning. During the online learning stage, we use the collected distribution data from garments in $G_{\rm train}$ to set an \emph{informed} prior for the MAB, rather than using the \emph{uninformed} initial prior of $\mathcal{N}(0.5, 1)$ for all actions.  We then apply the MAB with the informed prior for a new test garment from test dataset  $G_{\rm test}$. We compare two methods for setting the prior mean and standard deviation based on prior experience: \emph{All Garment} and \emph{Category}.  \textbf{All Garment} uses the empirical mean and standard deviation of coverage across all training garments for those corresponding actions. \textbf{Category} uses garments of only the same category as the test garment. 
\vspace{-10pt}
\subsection{Execution-Time Stopping Rules} \label{subsec:stopping_execution}
Given a novel garment, we execute the best action $a_{\rm{best}} := \rm{argmax}_{a \in \mathcal{A}} \mu(a)$ identified by the online learning procedure in Section~\ref{ssec:learning-adapt}; this is the action with the highest posterior mean coverage. However, due to aleatoric uncertainty arising from the garment state and air friction, there is significant variance in the outcome. To mitigate this uncertainty, the robot repeatedly flings the garment until the current coverage is above a learned threshold. We propose 3 methods for deciding when to stop. 
All methods use the model posterior over $a_{\rm{best}}$'s coverage, learned during training: $p(r \mid a_{\rm{best}}, \mathcal{D}_{\rm{train}}) = \mathcal{N}\left(\mu(a_{\rm{best}}), \sigma^2(a_{\rm{best}})\right)$, where $\mathcal{D}_{\rm{train}}$ is the dataset collected during training.  In this section, \emph{stopping rule} refers exclusively to execution-time.

The first two methods set a hard stopping threshold: if the coverage is above a threshold, the robot stops. As different garments may have different performance levels, for Stopping Rule 1, we set the threshold based on a z-score determined from $a_{\rm{best}}$'s posterior: $\mu(a_{\rm{best}}) + z\sigma(a_{\rm{best}})$ for some $z > 0$. For Stopping Rule 2, we set the threshold based on the one-step expected improvement calculated from the posterior $p(r \mid a_{\rm{best}}, \mathcal{D}_{\rm{train}})$ and stop if it falls below a given threshold.

Stopping Rule 3 accounts for a maximum flinging budget $B$. We decide to stop by estimating the expected improvement in coverage achievable during the remaining budget. We estimate this quantity using posterior sampling. Let $r_j^{\rm{exec}}$ be the coverage at step $j$ of execution. We draw $L$ sets of samples $\tilde{r}_{j+1, i}^{\rm{exec}}, \ldots, \tilde{r}_{B, i}^{\rm{exec}} \sim p(r \mid a_{\rm{best}}, \mathcal{D}_{\rm{train}})$, $i \in \{1, \ldots, L\}$, and approximate the expected improvement as:  
    $EI(a_{\rm{best}}) \approx \frac{1}{L}\sum_{i=1}^L \left[\max\{\tilde{r}_{j+1, i}^{\rm{exec}}, \ldots, \tilde{r}_{B, i}^{\rm{exec}}\} - r_j^{\rm{exec}}\right].$
One could straightforwardly penalize $EI(a_{\text{best}})$ based on the cost of performing each fling action. Without such a penalty, however, this more-optimistic stopping rule will tend to fling more times than the one-step lookahead rules.

\section{Experiments}
We use a UR5 robot arm to perform experiments across a total of 36 garments from 6 categories: T-shirts, long-sleeve shirts, towels, dresses, sweat pants, and jeans, with 30 garments in $G_{\rm train}$ and 6 garments in $G_{\rm test}$. See the project website for examples of garments in $G_{\rm test}$ at various smoothness levels. 
We calculate coverage via color thresholding on RGB images from a Logitech Brio 4K webcam.
At the start of each experiment batch with a given garment, the human supervisor manually fixes the grasp location.  The gripper never opens during the experiments. 
We use the following parameter ranges: $v^{23}_{max} \in [2 \text{m/s}, 3 \text{m/s}]$, $v^{34}_{max} \in [1 \text{m/s}, 3 \text{m/s}]$, $P_{3, y} \in [0.55 \text{m}, 0.7\text{m}]$, $P_{3, z} \in [0.4\text{m}, 0.55\text{m}]$, $\theta \in [-40\degree, 20\degree]$, $v_{\theta} \in [-1 \text{m/s}, 1 \text{m/s}]$, $a_{\theta} \in [-20 \text{m/s}^2, 20 \text{m/s}^2]$. Each fling cycle takes $\sim$45\,s.

To evaluate problem difficulty and our algorithm's effectiveness, we compare with five human subjects. 
Results suggest that the robot achieves similar performance to humans. Details of this experiment are shown in the project website.

\begin{figure}[t]
\center
\includegraphics[width=0.8\textwidth]{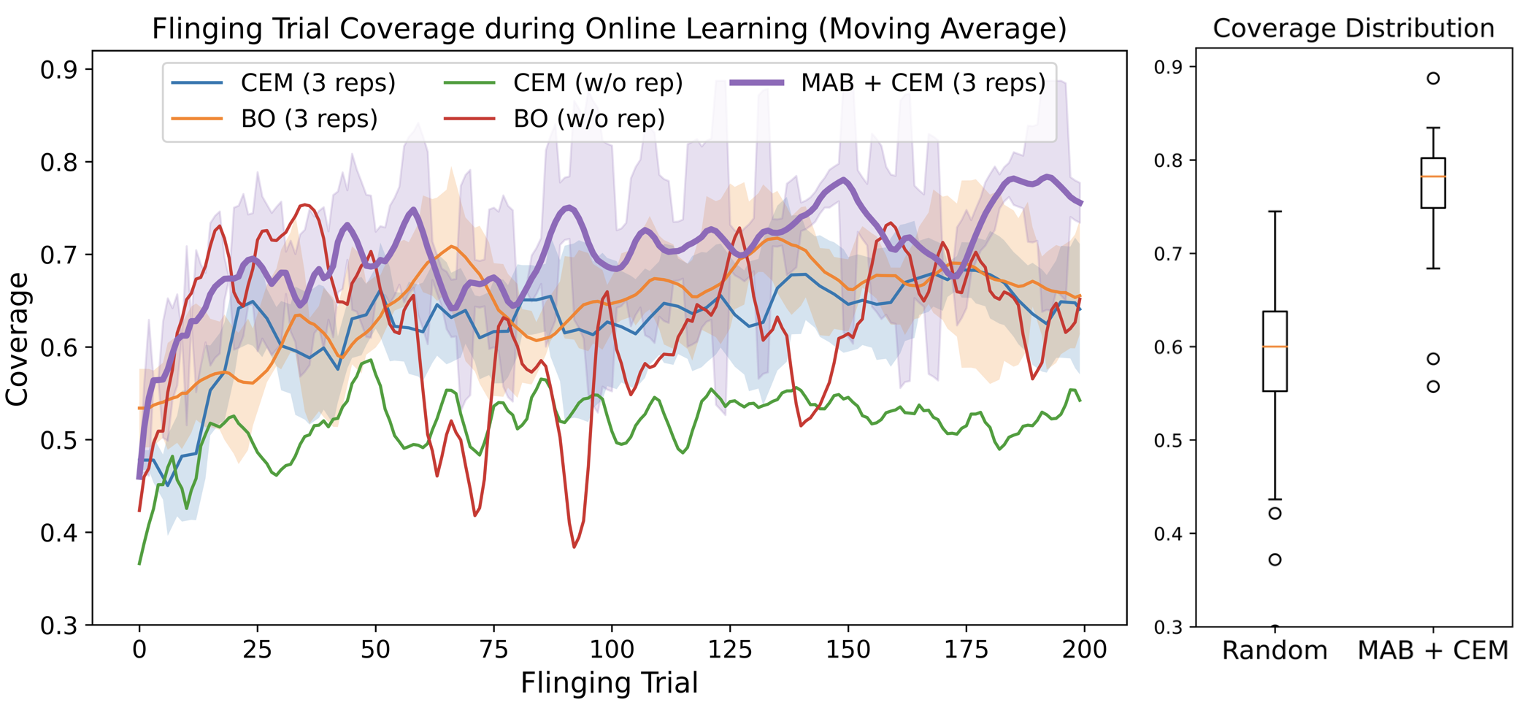}
\caption{Physical experiments showing the coverage of a T-shirt (fully-smoothed = 100\%) vs. number of fling trials for MAB with CEM refinement (MAB+CEM) compared with baselines. \textbf{Left:} Coverage achieved over time for each optimization algorithm, smoothed by moving window averaging (window size = 10). For CEM and BO with 3 repetitions per sampled action, the shaded region shows the standard deviation among the 3 trials for each action. MAB+CEM converges more quickly than either continuous optimization algorithm alone. \textbf{Right:} The post-convergence coverage distribution of MAB+CEM significantly outperforms random action selection. 
}
\vspace{-20pt}
\label{fig:method_comp}
\end{figure}

\vspace{-10pt}
\subsection{Comparison of Learning Methods}
\label{learning1}
We first study different learning methods. We use a T-shirt and compare the learning performance of MAB+CEM with 3 baselines on that T-shirt: 
CEM, Bayesian optimization (BO)~\cite{snoek2012practical}, and random actions. CEM and BO both perform continuous optimization without MAB, and BO uses a Gaussian Process model and the expected improvement acquisition function. The random baseline samples a random action from the continuous parameter range in each trial.

For MAB, we define the action space $\mathcal{A}$ via a uniform grid discretization of the trajectory parameters. To keep $|\mathcal{A}|$ small, we fix $\theta$, $v_{\theta}$, and $a_{\theta}$ at the centers of their ranges and vary the other 4 parameters, since we notice that $v^{23}_{max}$, $v^{34}_{max}$, $P_{3, y}$, and $P_{3, z}$ most strongly affect fling performance. We divide this 4-parameter space into a $2^4$-cell grid and use the resulting grid cell centers as bandit actions. 

For BO and CEM, we optimize over the entire continuous parameter range. Since BO and CEM are noise sensitive, we repeat each action 3 times, averaging the coverage values. A CEM iteration samples 5 actions and forms the elite from the top 3. We run BO for 70 iterations ($70\times3=210$ trials) and run CEM for 14 iterations ($14\times5\times3=210$ trials).  For MAB+CEM, we run MAB for 50 iterations and then CEM for 10 iterations ($50 + 10\times5\times3 = 200$ trials). We also compare BO and CEM without repetition, for 210 and 42 iterations respectively.

Results in Fig.~\ref{fig:method_comp} suggest that MAB+CEM outperforms CEM and BO with repetition. Due to the large aleatoric uncertainty in each fling trial, as can be seen from the shaded region and the box plot, BO and CEM without repetition perform the worst and are either unstable or fail to converge to a good action. 
\begin{figure}[!tbp]
  \begin{minipage}[t]{0.45\textwidth}
  \centering
    \includegraphics[height = 3.7cm]{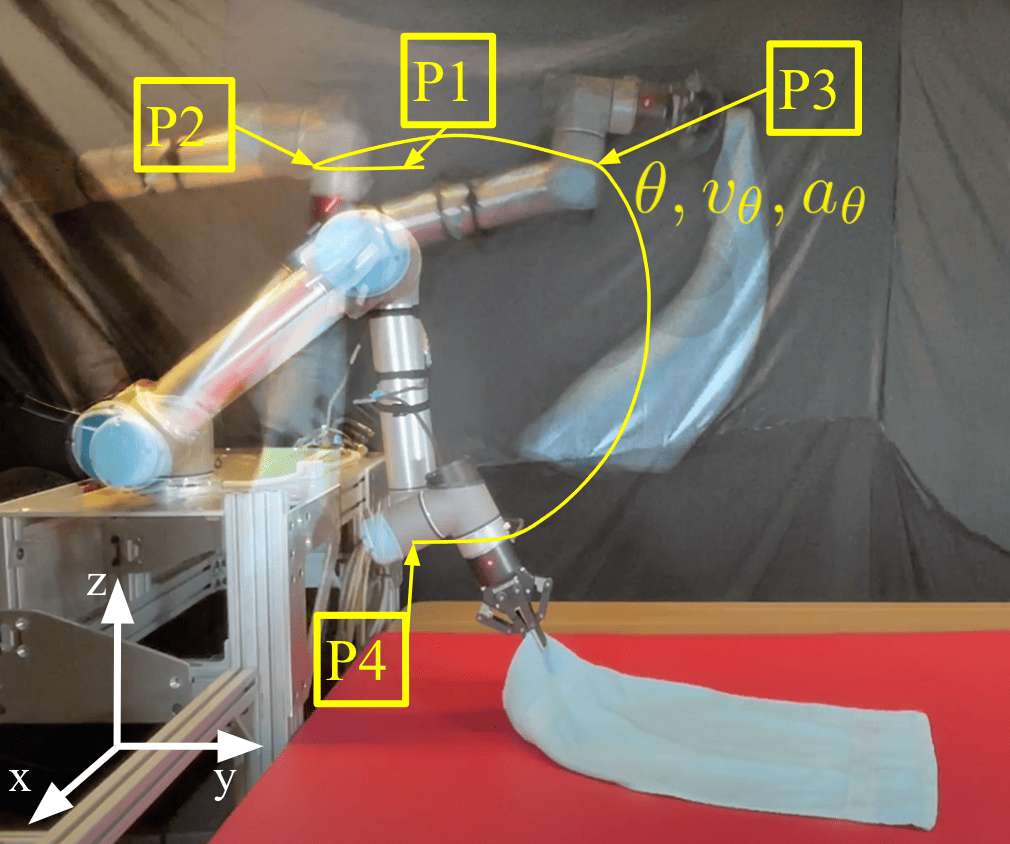}
    \caption{7-Dim trajectory parameterization. The robot end effector begins at point P1, moves to P2, moves (flings) to P3 at a high speed, and travels downwards to P4, bringing the garment to rest. $\theta$ is the angle of joint 3 at P3; $(v_{\theta}, a_{\theta})$ are the velocity and acceleration of joint 3 at P3. 
    }
    \label{fig:traj}
  \end{minipage}
  \hfill
  \begin{minipage}[t]{0.52\textwidth}
  \centering
    \includegraphics[width=1\textwidth]{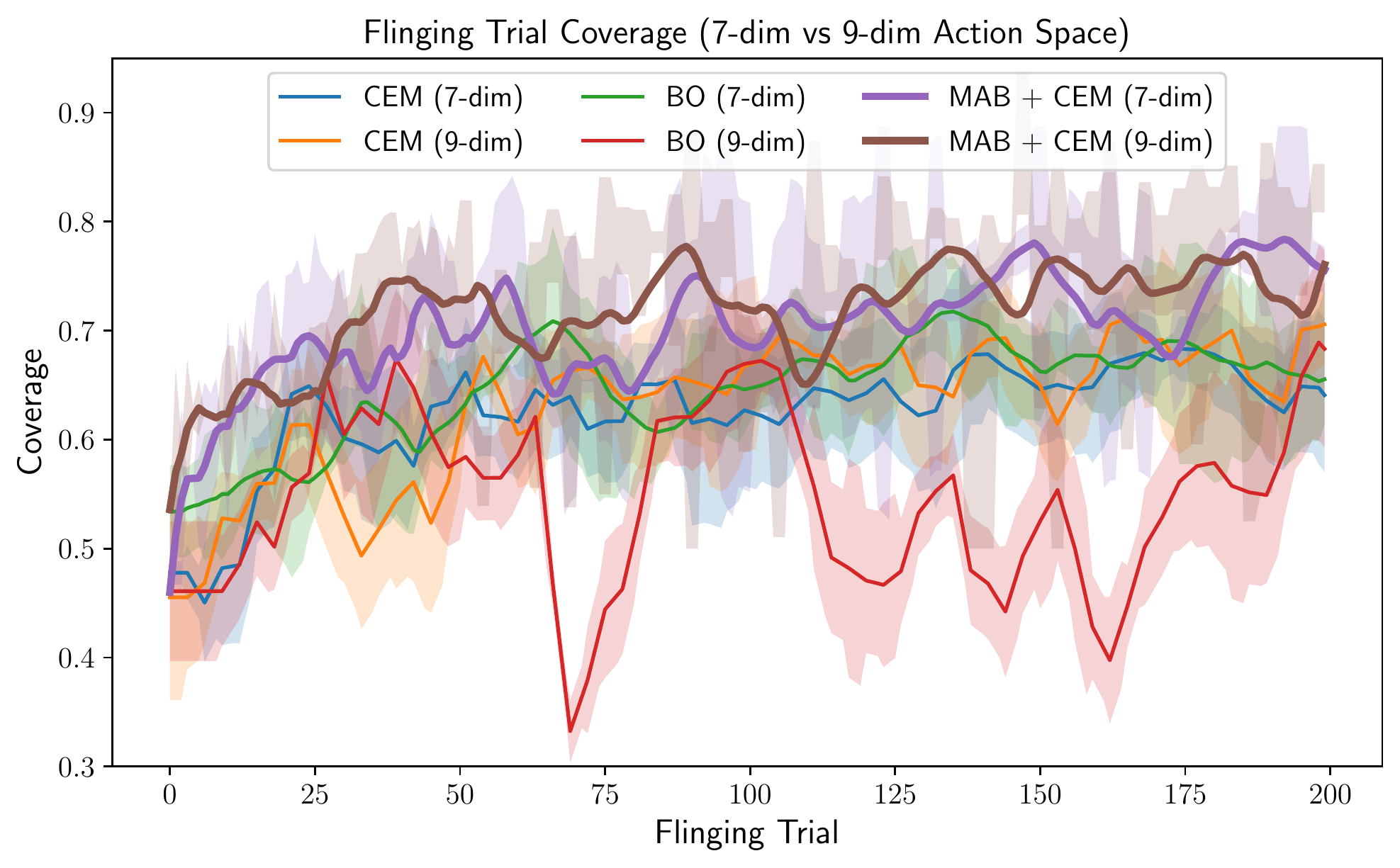}
    \caption{We conduct physical experiments to compare optimizing in 7 and 9-dimensional parameter spaces, where in the latter, the acceleration parameters are also learnable. Each action is repeated 3 times for CEM and BO. We find that 7 parameters are sufficient.
    }
    \label{fig:method_comp_79}
  \end{minipage}
  \vspace{-20pt}
\end{figure}

\vspace{-12pt}
\subsection{Comparison of Different Trajectory Parameterization Methods}
To study the trajectory parameterization, we also compare the learning performance of the 7D parameterization with a 9D parameterization on the same T-shirt used in Section~\ref{learning1}. The 9D parameter space consists of the original 7 parameters and two acceleration parameters: $a^{23}_{max}$ and $a^{34}_{max}$, corresponding to the maximum acceleration for all joints while traveling from P2 to P3 and from P3 to P4, respectively.
Results in Fig.~\ref{fig:method_comp_79} indicate that the 7D and 9D parameterizations exhibit similar performance for MAB+CEM, while the 9D parameterization makes BO less stable. MAB+CEM outperforms either CEM or BO alone in both parameterizations, suggesting generalizability of our proposed coarse-to-fine method. 
\vspace{-12pt}
\subsection{Results for Different Grasp Points}
\label{ssec:different_grasp_points}
\begin{figure}
  \centering
  \begin{minipage}[b]{0.22\textwidth}
    \includegraphics[width=\textwidth]{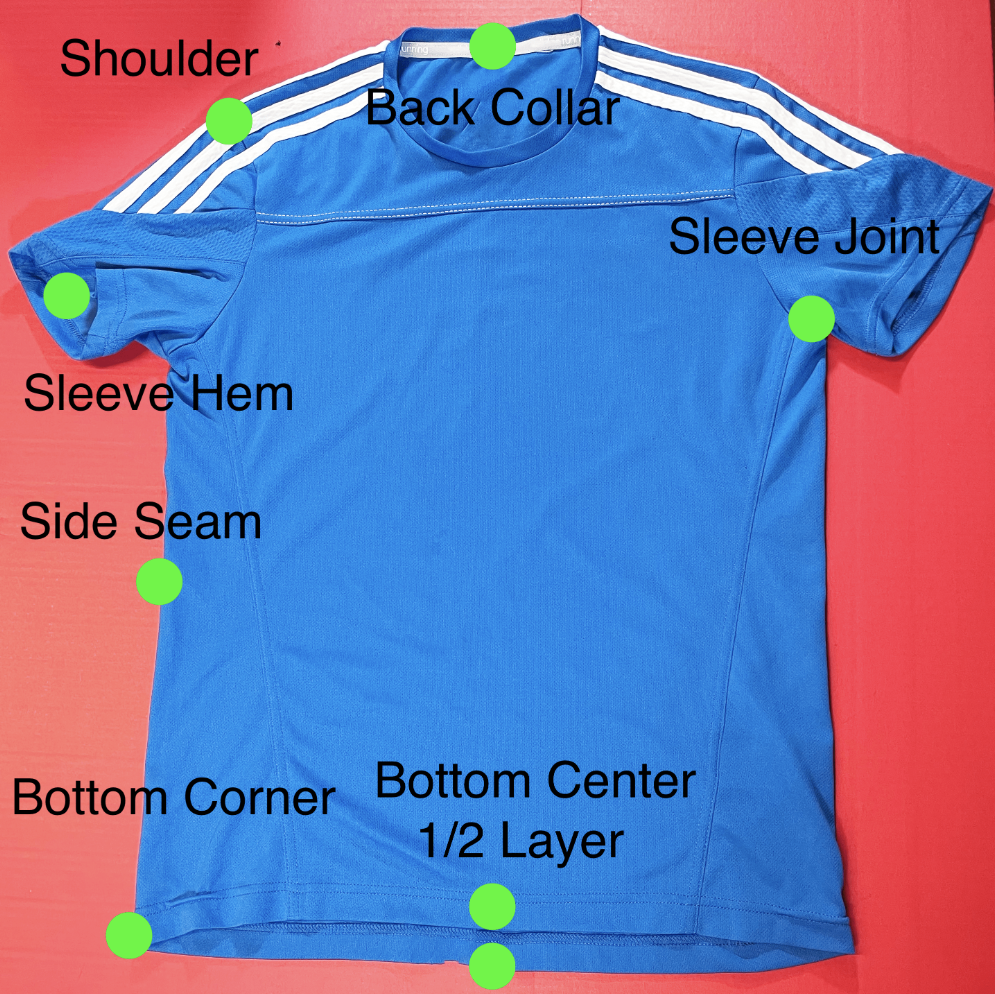}
    \caption{8 T-shirt grasp points. 
    }
    \label{fig:result-grasp_point}
  \end{minipage}
  \hfill
  \begin{minipage}[b]{0.75\textwidth}
   \captionsetup{type=table}
    \centering
    \scriptsize
    \begin{tabular}{@{}c@{\qquad}c@{\qquad}c@{}}
    \toprule
    Grasp Point             & Median Coverage & Max Coverage  \\ \midrule
    Bottom Center (1-layer) & 44\%            & 54\%          \\
    Side Seam               & 47\%            & 57\%          \\
    Bottom Corner           & 49\%            & 66\%          \\
    Sleeve Joint            & 50\%            & 65\%          \\
    Shoulder                & 52\%            & 66\%          \\
    Bottom Center (2-layer) & 55\%            & 74\%          \\
    Sleeve Hem              & 55\%            & 70\%          \\
    Back Collar                  & \textbf{61}\%   & \textbf{74}\% \\ \bottomrule
    \end{tabular}
    \caption{
      Average coverage at each grasp point by the median and best actions among the 16 bandit actions used for MAB learning. 
    }
      \label{tab:grasp_point}

  \end{minipage}
  \vspace{-20pt}
\end{figure}

The choice of grasping point can affect single-arm fling performance. 
We study the effectiveness of single-arm fling actions on a T-shirt with various grasp points:
the back collar, sleeve hem, sleeve joint, shoulder, side seam, bottom center (1 layer and 2 layers), and bottom corner (Fig.~\ref{fig:result-grasp_point}). We apply the MAB algorithm for each grasp point. 

\vspace{-10pt}

Table~\ref{tab:grasp_point} reports the average coverage of the median and best actions, and indicates that the back collar is the best grasp point, followed by the sleeve hem and bottom center (2-layer). Thus, 
we focus on grasping the back collar, while selecting analogous grasp points for other garment types (Section~\ref{subsec:workspace_definition}).

\subsection{Training and Testing Across Different Garments}


We use MAB+CEM to learn fling actions for different garments. For MAB, we set the threshold for the expected improvement (EI) to be 1.5\,\% because empirically, we find that 
the EI starts to decrease significantly more slowly past 1\,\%. 
We then refine via 2 iterations of CEM ($2\times5\times3 = 30$ trials). 

For each garment category, the training set $G_{\rm train}$ includes 5 garments. For each garment in $G_{\rm train}$, we run MAB with 16 actions for $N_{\rm train} = 50$ trials. We do not perform the CEM step, as the goal is to form a prior on those 16 actions which we use to accelerate the learning on test garments, instead of finding the best action on each training garment. 

\begin{figure*}[h!]
\center
\includegraphics[width=\textwidth]{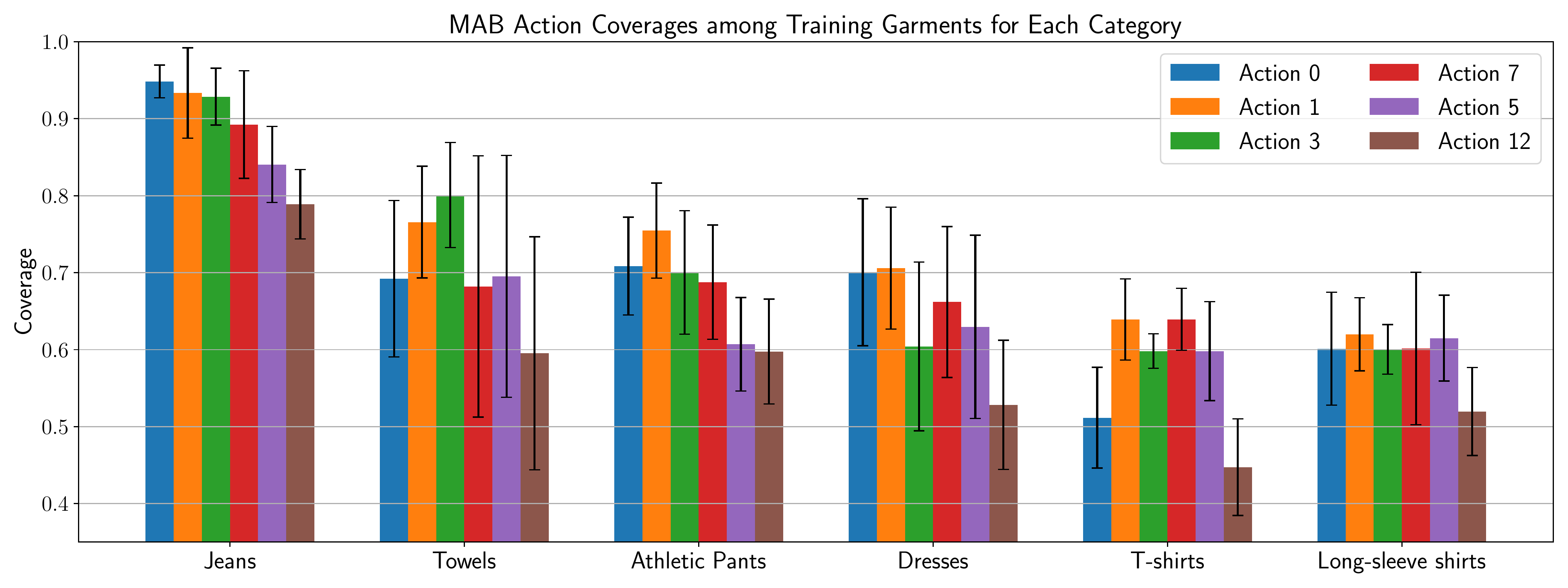}
\caption{From physical experiments, average coverage achieved for 6 of the 16 bandit actions among all garments in the training set $G_{\rm train}$, separated by garment category (mean $\pm$ 1 std). Each category contains 5 garments. The standard deviations indicate performance variability among different garments within a category. 
While some actions tend to outperform others, the best actions differ for different garment categories.  
}
\vspace{-20pt}
\label{fig:mab_prior}
\end{figure*}

Fig.~\ref{fig:mab_prior} shows the average coverage for 6 of the 16 bandit actions, separated by category. We see that within each garment type, different actions lead to significant differences in coverage. Moreover, the relative performance of various bandit actions is different in each category, suggesting that different actions are better suited for different garment categories.

\begin{table*}[t]
  \setlength\tabcolsep{5.0pt}
  \centering
 \begin{adjustbox}{max width=\textwidth}
  \begin{threeparttable}
  \begin{tabular}{@{}c@{~}||crrrl||crrrr@{}}

  \toprule
 Prior & Type & $\#$ & \multicolumn{1}{c}{MAB} & \multicolumn{1}{c}{+CEM} & p-val. & Type & $\#$ & \multicolumn{1}{c}{MAB} & \multicolumn{1}{c}{+CEM} & p-val.\\
    \midrule
 Uninf.    &       &24  & 94$\pm$4\%  & 94$\pm$1\% & 0.509 & & 32  & 93$\pm$6\%   & 93$\pm$7\%    & 0.594  \\
 All  & Jeans   & 9  & 92$\pm$1\% & 93$\pm$2\% & 0.003 & Towel & 16  & 93$\pm$6\%   & 93$\pm$7\%    & 0.594  \\
 Cat.  &         & \textbf{3}  & 94$\pm$4\%  & 94$\pm$1\%  & 0.509  &   & \textbf{12} & 90$\pm$6\% & 88$\pm$4\% & 0.759\\

    \midrule
 Uninf.      &      & 40  & 78$\pm$2\% & 78$\pm$2\% & 0.743 & \multirow{3}{2em}{Sweat Pants}  & 30 & 75$\pm$5\% & 78$\pm$4\%  & 0.011 \\ 
 All &  Dress & 30  & 79$\pm$3\% & 82$\pm$3\%  &  0.003  & & 20 & 77$\pm$3\%   & 80$\pm$3\%    & 0.016   \\ 
 Cat.   & & \textbf{6}  & 80$\pm$1\% & 81$\pm$2\%  &  0.035 &   & \textbf{13} & 77$\pm$3\%   & 80$\pm$3\%    & 0.016   \\

  \midrule
 Uninf.       &     & 32  & 72$\pm$6\% & 73$\pm$3\%  & 0.257 &   \multirow{3}{2em}{Long-sleeve Shirt}   & 48 & 57$\pm$5\% & 60$\pm$5\% & 0.060 \\
 All  & T-shirt & 19  & 73$\pm$6\%   & 75$\pm$4\%    & 0.115   &     & 37 & 54$\pm$6\% & 61$\pm$5\% &  4e-4\\
 Cat.  &   & \textbf{7}  & 73$\pm$6\%   & 75$\pm$4\%    & 0.115   &   & \textbf{14} & 57$\pm$4\% & 60$\pm$4\% & 0.018 \\

  \toprule
  \end{tabular}
\end{threeparttable}
  \end{adjustbox}
  \caption{Physical experiments with 36 garments demonstrating the efficiency of learning fling actions on test garments using category priors.
  Under each MAB prior, we report the number (\#) of bandit trials that MAB takes to reach a stopping threshold (EI $<$ 1.5\%), and columns 4, 5, and 9, 10 are the average coverage (mean $\pm$ std) achieved by the best MAB-identified action (prior to CEM) and by MAB+CEM, where we execute each action 20 times. We report identical coverage numbers for MAB+CEM instances in which MAB identifies the same best action, since the CEM procedure is the same. 
  We compare the \emph{Uninformed} prior (Uninf.) to the \emph{All-Garment} (All) and \emph{Category} (Cat.) priors, for which the decrease in MAB trials indicates high garment category informativeness. 
  The p-values are from t-tests comparing whether the mean coverage of the action identified by MAB+CEM significantly improves on the MAB-identified action; 
  p-values compare the difference in sample means with standard errors, which account for the sample size (20). Thus, while there is large aleatoric uncertainty, as seen from the large standard deviations among the 20 repetitions, the single-digit percentage increase in mean can be statistically significant. 
  }
  \vspace*{-30pt}
  \label{tab:prior_comp}
\end{table*}

Using the data from $G_{\rm train}$ to inform the MAB prior, we evaluate performance on the garments in $G_{\rm test}$, which contains one new garment from each category. We compare MAB performance with 3 priors: a) the original prior (i.e., ``Uninformed''), b) a prior calculated using all the garments in $G_{\rm train}$ (``All-Garment''), and c) a prior calculated only using garments in $G_{\rm train}$ of the same category as the test garment (``Category''). The informed priors (b and c) set the prior mean and standard deviation for each bandit action using the empirical means and standard deviations from training. Table~\ref{tab:prior_comp} reports results over 20 repetitions. While the different priors do not significantly affect the final coverage achieved by the best-identified action, the results suggest that the ``Category'' prior can significantly accelerate learning. In particular, with a more informative prior based on $G_{\rm train}$, a new garment's EI falls below the termination threshold in 57--87\,\% fewer MAB iterations than when starting from an uninformed prior.

From Table~\ref{tab:prior_comp}, we also see that CEM refinement can usually increase performance by several percentage points over the best MAB-identified action, with the exception of the jeans and towel, where MAB already identifies high-coverage actions. The p-values test whether the improvements in mean coverage are statistically significant, given the large aleatoric uncertainty of each individual fling trial (whose coverages have standard deviation up to 7\,\%). 
We conclude that within 10 MAB iterations (10 trials) and 2 CEM iterations (30 trials), the robot can efficiently learn a fling action that achieves 60--94\,\% coverage for a new garment in under 30 minutes. For comparison, using two arms, Flingbot~\cite{ha2021flingbot} achieves 79--93\,\% coverage on towels and 93--94\,\% coverage on T-shirts.
\vspace{-10pt}
\subsection{Stopping Time During Execution} 
\label{exp:stopping}
\begin{figure*}[t]
\center
\includegraphics[width=0.32\textwidth]{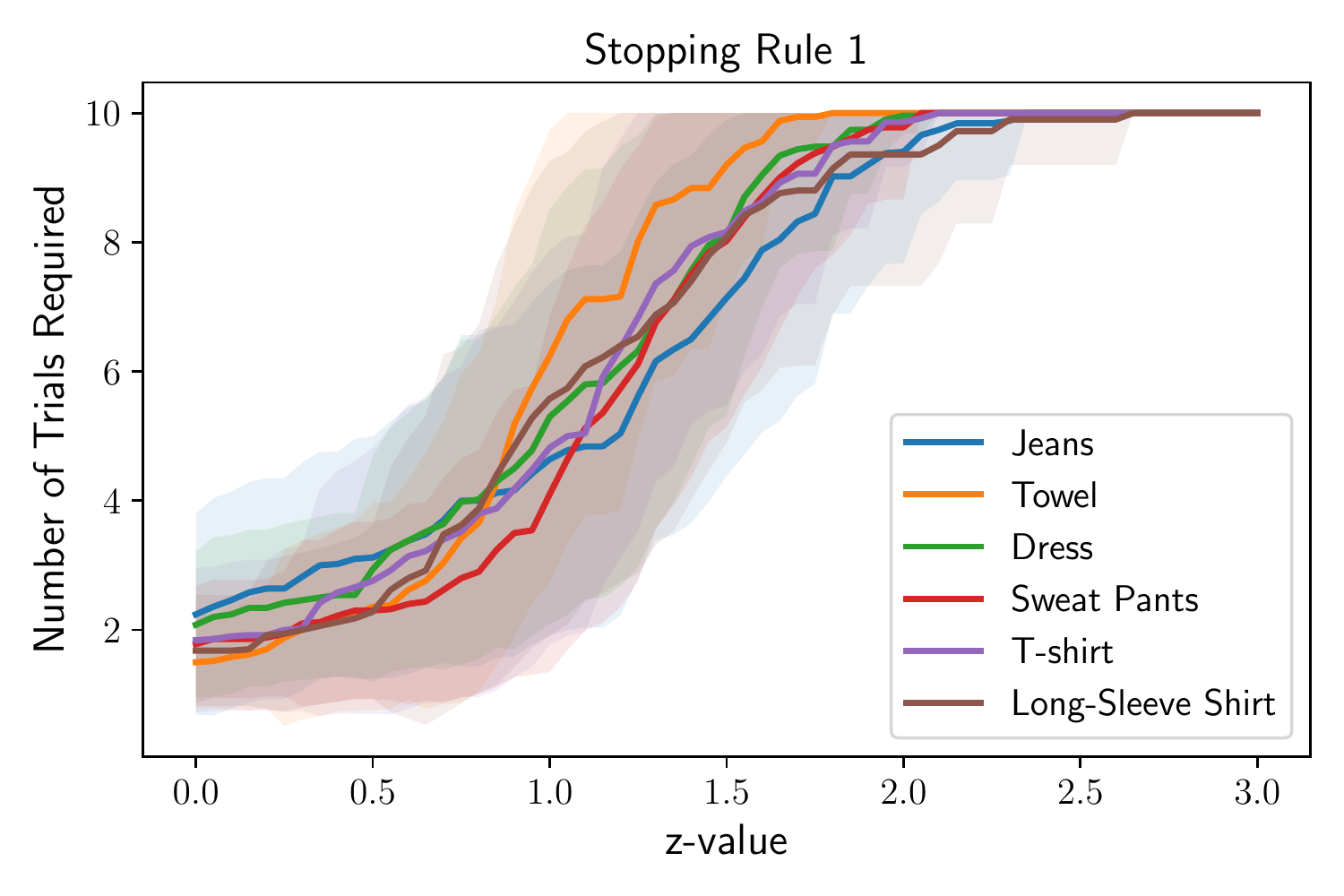}
\includegraphics[width=0.32\textwidth]{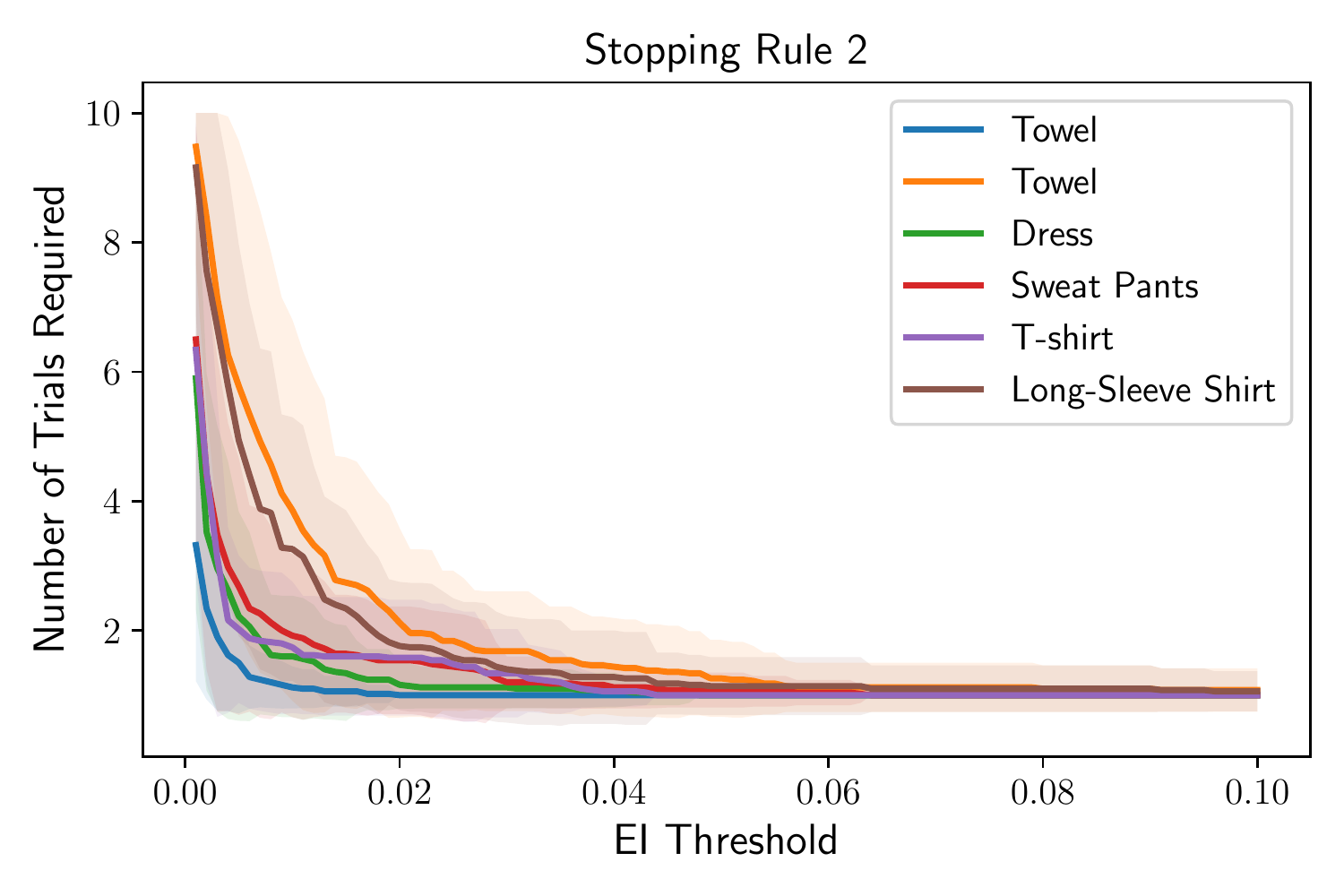}
\includegraphics[width=0.32\textwidth]{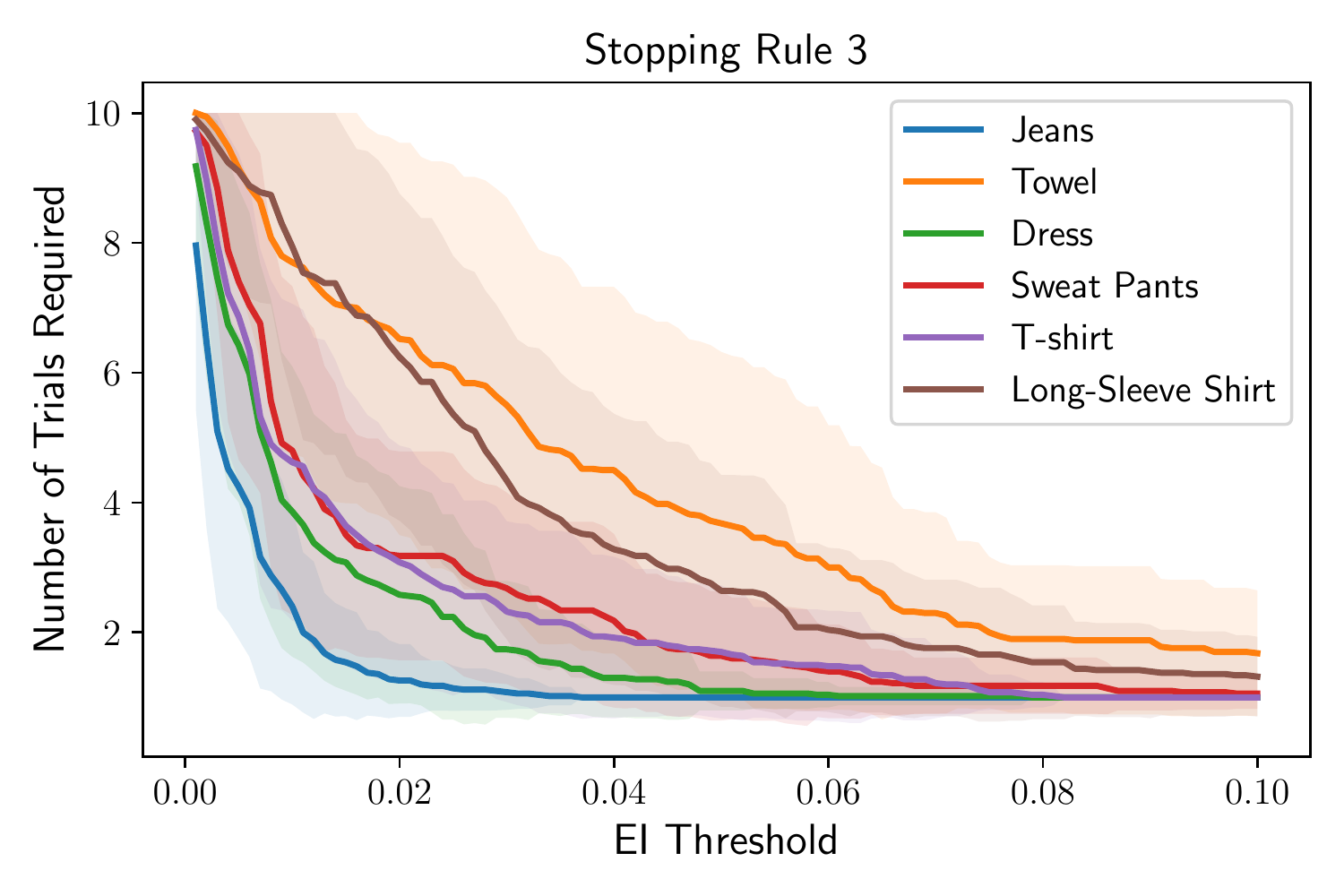}
\caption{
Number of trials under each execution-time stopping rule as a function of the termination threshold. \textbf{Left}: stopping when the coverage exceeds one standard deviation above the mean. \textbf{Middle}: stopping when the one-step expected improvement falls below a threshold. \textbf{Right}: stopping when the expected improvement over the remaining trial budget falls below a threshold, using a total budget of 10 trials. With repeated fling trials and optimal stopping, the robot can achieve better-than-average results during execution, especially for garments with large aleatoric uncertainty. 
}
\vspace{-15pt}
\label{fig:result-stopping}
\end{figure*}

To compare the 3 stopping rules proposed in Section \ref{subsec:stopping_execution} and study the trade-off between the number of flinging trials and performance, we apply each stopping rule with a range of thresholds by post-processing the execution-time results of the best action found for each garment. For each method, we bootstrap the experiment results and empirically estimate the stopping-time distribution under various stopping thresholds. From the results in Fig.~\ref{fig:result-stopping}, we can see that it takes on average 4 flings to achieve a coverage that is 1 standard deviation above the mean or that results in a one-step EI below 1\,\% absolute coverage. Stopping based on EI with a budget leads to more optimistic EI and significantly more trials. In practice, a heuristically-estimated cost for each additional fling can counter the tendency of this method to fling too many times.



\vspace{-10pt}
\section{Conclusion} 
\label{sec:conclusion}
We consider single-arm dynamic flinging to smooth garments. We learn fling motions directly in the real world to address complex, difficult-to-model garments such as long-sleeve shirts and dresses. We propose stopping rules during training and execution time. Results suggest that we can learn effective fling actions that achieve up to 94\,\% average coverage in under 30 minutes.
In future work, we will enhance the current system with quasistatic pick-and-place actions to fine-tune smoothing~\cite{lerrel_2020,seita_fabrics_2020} and explore evaluation metrics that judge the quality of dynamic motions by the ease of subsequent quasistatic fine-tuning. 

\section*{Acknowledgements}
{\footnotesize
This research was performed at the AUTOLAB at UC Berkeley in affiliation with the Berkeley AI Research Lab, the CITRIS ``People and Robots'' (CPAR) Initiative, and the Real-Time Intelligent Secure Execution
(RISE) Lab. The authors were supported in part by donations from Toyota Research Institute, Google, Siemens, Autodesk, and by
equipment grants from PhotoNeo, NVidia, and Intuitive Surgical.}


\appto{\bibsetup}{\sloppy}
{\small
\printbibliography 
}

\end{document}